\documentclass[letterpaper,twocolumn,fleqn]{article} 

\newcommand{\etal}{\textit{et al.}}

\usepackage{ist}
\usepackage{amsmath,graphicx}
\usepackage[dvipsnames]{xcolor}
\usepackage{tablefootnote}
\usepackage{booktabs}
\usepackage{multirow}
\usepackage{adjustbox}
\usepackage{siunitx}
\usepackage{float}
\usepackage{hyperref}
\hypersetup{
    colorlinks=true,
    linkcolor=blue,
    filecolor=magenta,
    urlcolor=cyan,
}
\usepackage{amssymb}
\usepackage{pifont}

\title{How Much Depth Information can Radar Contribute to a Depth Estimation Model?}

\author{ 
Chen-Chou Lo and Patrick Vandewalle; \\
EAVISE, PSI, KU Leuven; \\ 
Sint-Katelijne-Waver, Belgium.
}

\date{} 

\begin{document} 

\maketitle 

\thispagestyle{empty} 


\begin{abstract}
\vspace{1mm}
Recently, several works have proposed fusing radar data as an additional perceptual signal into monocular depth estimation models because radar data is robust against varying light and weather conditions.
Although improved performances were reported in prior works, it is still hard to tell how much depth information radar can contribute to a depth estimation model.
In this paper, we propose radar inference and supervision experiments to investigate the intrinsic depth potential of radar data using state-of-the-art depth estimation models on the nuScenes dataset.
In the inference experiment, the model predicts depth by taking only radar as input to demonstrate the inference capability using radar data. In the supervision experiment, a monocular depth estimation model is trained under radar supervision to show the intrinsic depth information that radar can contribute. 
Our experiments demonstrate that the model using only sparse radar as input can detect the shape of surroundings to a certain extent in the predicted depth.
Furthermore, the monocular depth estimation model supervised by preprocessed radar achieves a good performance compared to the baseline model trained with sparse lidar supervision. 
\end{abstract}


\section{Introduction}
\label{sec:intro}
\vspace{1mm}
Depth estimation plays an essential role as a fundamental piece of information for applications like 3D object detection and 3D reconstruction. Thanks to the development of deep neural networks in recent years, researchers have proposed many monocular and stereo depth estimation algorithms~\cite{PSMNet,GA-Net,eigen_1,eigen_2,DORN,bts} with significant improvement. However, for such camera-based methods, monocular depth estimation is ill-defined in the sense that many scenes could project to the same 2D image. Stereo depth estimation is sensitive to environmental lighting and texture conditions. Consequently, many works~\cite{MSG-CHN,guide_uncertainty,s2d} have been proposed to leverage additional depth information from lidar data as guidance to compensate for the ill-defined depth perception of camera features. This results in a more robust and accurate performance.

Since the projected depth from lidar sensors is very accurate and high-end lidar could result in a relatively high resolution projected depth, lidar has been the most commonly used depth sensor for guiding camera-based depth estimation models. Although lidar contributes a lot of highly accurate depth information about the surroundings, it is also notorious for its high cost and sensitivity to weather conditions. On the contrary, radar is known for its robustness and reliability against extreme weather, and it is much cheaper than lidar sensors. Accordingly, several researchers are now fusing radar sensor data as further guidance into camera-based depth estimation models~\cite{MDE_radar,DORN_radar,RC_PDA,MDE_multitask,RVMDE} after the release of the large autonomous driving dataset nuScenes~\cite{nuScenes} which includes monocular camera image, radar, and lidar data.

Existing multi-modal depth estimation methods that integrate sparse radar with camera images are based on the same structure of models that make use of camera images and sparse lidar. However, these lidar-fusion models are dedicated to extracting features from lidar data, and inferred depth from lidar features and radar features are quite different. These proposed radar-fusion works showed that the results improved after modifying parts of the original architectures and fusing radar features as additional guidance. However, unlike other data sources such as camera images or lidar data that carry rich accurate depth information, radar data is not only sparse but also noisy and view-limited. Still, it is hard to tell to what extent the radar can contribute to a depth estimation model. 

In this work, our goal is not to propose novel model architectures or radar preprocessing methods. 
Instead, we investigate how much depth information radar data can infer and contribute to a depth estimation model based on state-of-the-art models and radar preprocessing methods. We conduct two sets of experiments: firstly, to predict depth by using only radar as an input feature to demonstrate the inference potential using radar data. 
Secondly, to train a monocular depth estimation model with radar supervision instead of lidar supervision, showing the intrinsic depth information that radar contributes to models taking monocular image input. 
Through the experiments, we show that (1) the shape of surroundings is captured in the output prediction by a depth estimation model with only radar input, and (2) that a monocular depth estimation model trained with radar supervision can estimate the depth map to a good extent. 
The results agree with the positive results reported by prior works that depth estimation models can learn useful depth representation from radar input with lidar supervision.
Additionally, the results from radar supervision experiment also show the potential of radar data as a guidance signal in depth estimation tasks and ease the demand for lidar data. 
To the best of our knowledge, there is no previous work investigating radar on these two aspects.



\begin{figure*}
  \includegraphics[width=\textwidth,height=\textheight,keepaspectratio]{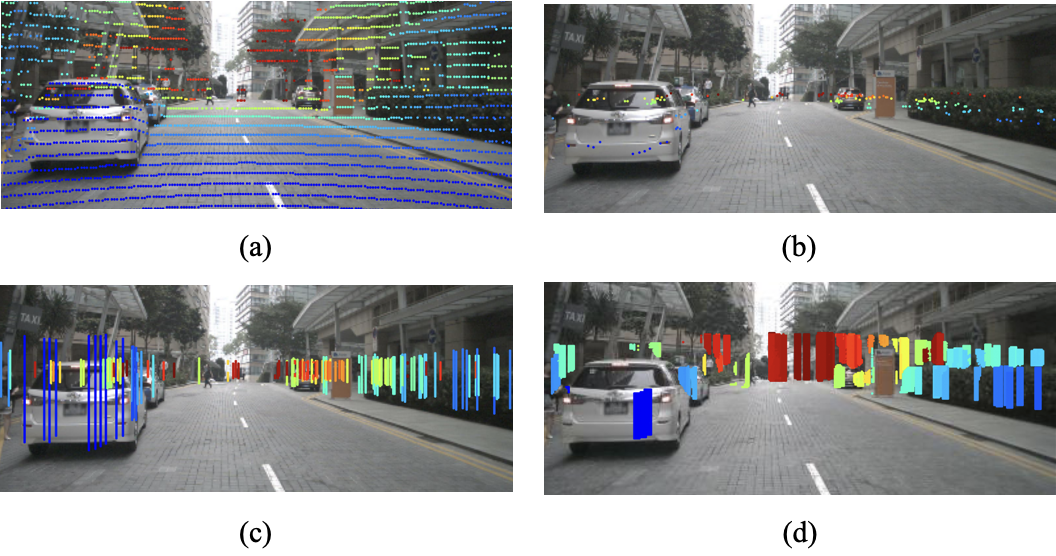}
  \caption{Sample images from nuScenes. (a) An image with 1 sweep of sparse lidar projection; (b) 5 sweeps of raw sparse radar projection; (c) 5 sweeps of height-extended radar projection; (d) MER channel with RC-PDA $\geq 0.5$. All the point sizes are dilated for better visualization.}
  \label{fig:sample}
\end{figure*}



\section{Related Work}
\label{sec:related_work}
\vspace{1mm}
Many monocular depth estimation models have been proposed these years by exploiting the power of neural networks~\cite{eigen_1} with different architectures~\cite{bts, BANet} and loss functions~\cite{DORN}. Some works even involve additional semantic learning~\cite{zhu2020edge}, or self-supervision learning~\cite{monodepth1, monodepth2} that leverages frames from the time domain.
In this section, we focus on the related depth estimation works that are based on multi-modality.

\subsection{Depth Estimation using Camera and Lidar}
\vspace{1mm}
Many works have proposed using lidar as the additional guidance signal since integrating lidar into a monocular depth estimation model can significantly improve the overall performance. Ma \etal~\cite{s2d} first proposed an early-fusion approach to concatenate sparse lidar data and monocular camera images to form the input features. In addition, a CNN encoder-decoder architecture was used to estimate a dense prediction based on the concatenated input.
In contrast, Jaritz \etal~\cite{SparseAD} used late fusion for lidar and camera image feature integration and multi-task learning to improve the performance of depth estimation. 
Next to the feature map fusion strategy, Vangansbeke \etal~\cite{guide_uncertainty} used two separate branches to output prediction based on camera image and lidar input, and the two estimated depth features are merged into a final estimation via confidence maps integration. 
Li \etal~\cite{MSG-CHN} investigated supervision from multi-scale ground-truth with a cascade hourglass network to leverage the structure from low to high resolution. 

\subsection{Depth Estimation using Camera and Radar}
\vspace{1mm}
Meanwhile, researchers have also engaged in fusing sparse radar data into depth estimation models for radar's robustness and reliability in different conditions. 
Lin \etal~\cite{MDE_radar} firstly conducted comprehensive experiments based on different fusion approaches and proposed a two-stage prediction method to tackle the noise in radar data. 
In our earlier work~\cite{DORN_radar}, we proposed a radar preprocessing method by projecting a given radar point to a pre-defined real-world height and mapping onto camera coordinates to tackle the limited vertical field of view and sparsity characteristics of radar data. Furthermore, a deep ordinal regression model, DORN, is used as the backbone model to estimate the dense depth prediction.
Lee \etal~\cite{MDE_multitask} proposed multi-task learning and built upon the architecture from~\cite{MDE_radar} with additional detection heads for semantic segmentation and 2D object detection to improve the performance of depth estimation. 
Long \etal~\cite{RC_PDA} developed a two-stage algorithm that resolves some of the uncertainty of projected radar to densify projected radar depth at the first stage. A standard depth estimation approach is then used to estimate dense depth based on fused radar and image data.

All the existing depth estimation models using camera and radar data have reported promising results indicating that integrating radar can improve the accuracy of a camera-based depth estimation model. Our goal in this paper is not to introduce a new method for integrating radar and monocular depth estimation but rather to explore the bounds of how much depth information radar data can contribute.



\begin{figure*}
  \includegraphics[width=\textwidth,height=\textheight,keepaspectratio]{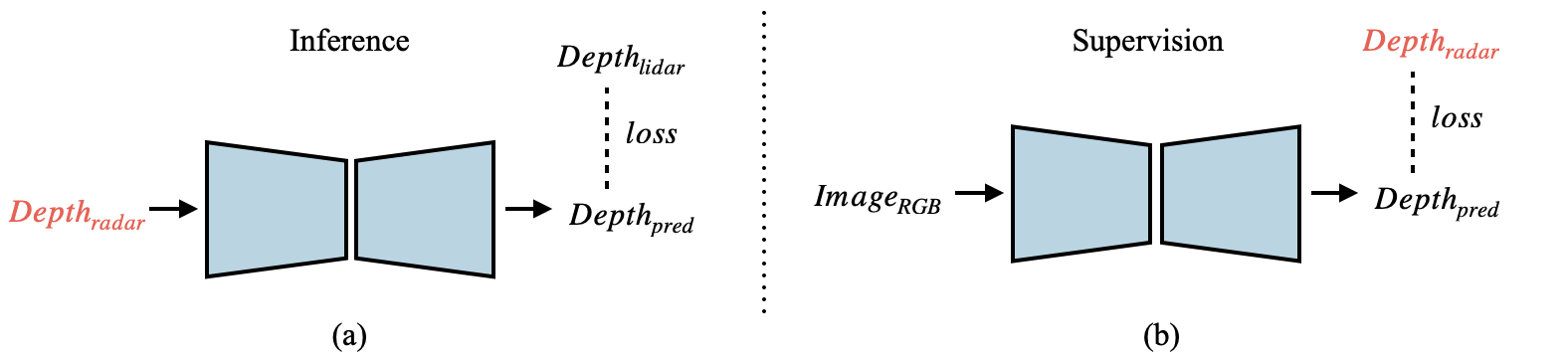}
  \caption{Illustration of the proposed (a) radar inference experiment; (b) radar supervision experiment.}
  \label{fig:exp}
\end{figure*}


\section{Method}
\label{sec:method}
\vspace{1mm}
In this section, we first introduce the different radar formats we use in our experiments. Then, the model architectures used for both experiments are described in detail in the following subsection.

\subsection{Radar Data}
\vspace{1mm}
{\bf Raw Radar.}
Although radar is a low-cost and robust sensor, a few characteristics bring disadvantages to radar as a depth guidance signal. As described in detail in~\cite{MDE_radar, DORN_radar, RC_PDA}, the main disadvantages are sparseness, noisy measurements, and limited vertical field of view. For tackling sparseness and noise issues, Lin \etal~\cite{MDE_radar} accumulated raw radar points from multiple frames and used prediction from the first stage to do filtering on the raw noisy sparse radar. In this work, we directly used multiple-frame raw radar as our raw radar features since we have no first-stage estimation from camera images for filtering. 

\vspace{1mm}
\noindent
{\bf Height-extended Radar.}
Besides raw radar data, we proposed in our earlier work 
~\cite{DORN_radar} the height-extended radar, extending the radar data from a point to a fixed height range of 0.25m to 2m in the world coordinates. By extending the height of radar points, the sparseness and limited view issues are mitigated, which helps gain depth estimation performance.
Therefore, we used height-extended radar data as one of the radar data in our experiments. 

\vspace{1mm}
\noindent
{\bf Multi-channel Enhanced Radar (MER).}
In addition to raw and height-extended radar data, Long \etal~\cite{RC_PDA} proposed a radar-camera pixel depth association (RC-PDA) method to generate MER at the first stage of their methods. The assumption is that radar can measure precise depth but the projected point on a camera view is shifted on height and width coordinates. Instead of finding the correct projected position, Long eased the problem by determining which positions in the neighboring region of the radar point have the same depth. A U-Net structure was used to train to estimate the similarity probability and confidence within each neighboring region of corrected radar depth under the supervision of ground truth lidar. The output predicted depth was formed into MER, and MER has six channels which refer to different confidence, or radar-camera pixel depth association (RC-PDA), as 0.5, 0.6, 0.7, 0.8, 0.9, and 0.95, respectively. In our experiments, we use the MER channel with RC-PDA $\geq 0.5$. 

A visualization of a sample image of nuScenes with projected raw sparse lidar, raw radar, height-extended radar, and MER channel with RC-PDA $\geq 0.5$ is shown in Fig.\ \ref{fig:sample}.


\newcommand{\ra}[1]{\renewcommand{\arraystretch}{#1}}
\begin{table*}
\caption{Table 1: Evaluation results for radar inference experiments with different methods and input radar. Note that the ground truth sparse lidar is used as the supervision signal in this experiment. CAP refers to the maximum depth range in meters.}
\label{tab:radar_inference}
\centering
\vspace{5mm}
\begin{tabular}{ccccccc}
\hline\hline
\multicolumn{1}{l}{model} & input radar & CAP & $\delta$$_1$ $\uparrow$ & $\delta$$_2$ $\uparrow$ & RMSE $\downarrow$ & AbsRel $\downarrow$ \\ 
\hline
& Raw radar & 80 & 0.716 & 0.774 & 7.817 & 0.260 \\
{DORN$_{radar}$} & Height-extended radar & 80 & {\bf 0.763} & 0.844 & {\bf 6.582} & 0.232 \\
& MER (RC-PDA $\geq 0.5$) & 80 & 0.736 & {\bf 0.902} & 7.781 & {\bf 0.227} \\
\hline
& Raw radar & 80 & 0.714 & 0.768 & 8.151 & 0.247 \\
{S2D$_{radar}$} & Height-extended radar & 80 & 0.783 & 0.865 & {\bf 6.404} & 0.220 \\
& MER (RC-PDA $\geq 0.5$) & 80 & {\bf 0.801} & {\bf 0.890} & 7.290 & {\bf 0.155} \\ 
\hline\hline
\end{tabular}
\end{table*}



\begin{table*}
\caption{Table 2: Evaluation results for radar supervision experiments. We use the RGB branch module of S2D$_{RGB}$ in this experiment. Note that the result of sparse-lidar is the RGB baseline result from~\cite{MDE_radar}. CAP refers to the maximum depth range in meters.} 
\label{tab:radar_supervision}
\centering
\vspace{5mm}
\begin{tabular}{cccccccc} 
\hline\hline
\multicolumn{1}{c}{supervision signal} & CAP & $\delta$$_1$ $\uparrow$ & $\delta$$_2$ $\uparrow$ & $\delta$$_3$ $\uparrow$ & RMSE $\downarrow$ & AbsRel $\downarrow$ \\
\hline
Sparse lidar~\cite{MDE_radar} & 80 & 0.862 & 0.948 & 0.976 & 5.613 & 0.126 \\
\hline
Raw radar & 80 & 0.292 & 0.522 & 0.644 & 18.995 & 0.707 \\
Height-extended radar & 80 & 0.602 & 0.778 & 0.846 & 12.511 & {\bf 0.288} \\
MER (RC-PDA $\geq 0.5$) & 80 & {\bf 0.605} & {\bf 0.849} & {\bf 0.920} & {\bf 8.837} & 0.302 \\
\hline
Raw radar & 50 & 0.321 & 0.552 & 0.672 & 13.996 & 0.600 \\
Height-extended radar & 50 & 0.609 & 0.784 & 0.852 & 10.661 & {\bf 0.280} \\
MER (RC-PDA $\geq 0.5$) & 50 & {\bf 0.612} & {\bf 0.8467} & {\bf 0.927} & {\bf 7.091} & 0.294 \\
\hline\hline
\end{tabular}
\end{table*}


\subsection{Architectures}
\label{subsec:architecture}
\vspace{1mm}
To investigate how much depth information radar can contribute, we conduct two types of experiments: a radar inference experiment and a radar supervision experiment. The input, output, and supervision signals in both experiments are illustrated in Fig.\ \ref{fig:exp}.

\subsubsection{Depth Estimation with only Radar Input}
\vspace{1mm}
One way to demonstrate the intrinsic depth capacity of sparse radar is to train a model that takes radar as the only input and is supervised by sparse lidar, as shown in Fig.\ \ref{fig:exp} (a). Thus, we make use of the depth branch from two state-of-the-art works. The depth branch module, termed DORN$_{radar}$, in~\cite{DORN_radar} is a simplified modification from DORN~\cite{DORN}. ResNet-26~\cite{ResNet} was used as the sparse radar feature extractor, and two $1\times 1$ convolutional layers were concatenated after the ResNet module. After the feature map extraction, a few deconvolutional layers were used for upsampling the feature map to the desired shape. The depth branch module, termed S2D$_{radar}$, in~\cite{MDE_radar} used a similar architecture as the model proposed in S2D~\cite{s2d}, and the input channel of the model was changed from 4 to 1 to fit the shape of input sparse radar depth. For the decoder part, bilinear upsampling operations were used in the upsampling layer. We use both models in this radar inference experiment to compare the effectiveness of model architectures.

\subsubsection{Monocular Depth Estimation with Radar Supervision}
\vspace{1mm}
In the second experiment, to show the intrinsic capacity of sparse radar to contribute depth information, we treat radar as a supervision feature to train a monocular depth estimation model as shown in Fig.\ \ref{fig:exp} (b). In this radar supervision experiment, we use the RGB branch in~\cite{MDE_radar} termed S2D$_{RGB}$. ResNet-18 was used as the RGB dense feature extractor. The whole structure is an encoder-decoder model that uses convolutional layers to encode the input RGB feature into a latent representation and bilinear upsampling layers to decode the representation into an output prediction. 

The RGB branch in DORN$_{radar}$ has the same structure as the original DORN. The input monocular camera image first goes through a pre-trained ResNet-101 to have its feature map extracted. A scene understanding module consisting of ASPP~\cite{ASPP} layers and image encoder layers is applied to further exploit the information within the feature map. Finally, some deconvolutional layers are used as upsampling layers to output the final prediction. 

\subsection{Loss Functions}
\vspace{1mm}
Since the loss functions used in the two works are different, we introduce both losses in this section. In the S2D model, the $\mathcal{L}_{L1}$ loss is used as the training criteria:

\begin{equation}
\label{eq:L1}
\mathcal{L}_{L1} = \frac{1}{WH} \sum_{w=0}^{W-1}\sum_{h=0}^{H-1} |y_{pred}(w,h)-y_{target}(w,h)|
\end{equation}

\noindent 
where ${(w,h)}$ is the pixel location while $y_{pred}$ and $y_{target}$ are the output prediction of the model and target depth, respectively.

On the other hand, DORN uses ordinal loss as the training criterium, which turns the regression problem into a classification problem. The ordinal loss $\mathcal{L}_{ordinal}$ is defined as the average of pixel-wise ordinal loss $\Psi(w,h,P)$ over the entire prediction:

\begin{equation}
\label{eq:Lord}
\mathcal{L}_{ordinal} = - \frac{1}{WH} \sum_{w=0}^{W-1}\sum_{h=0}^{H-1} \Psi(w,h,P)
\end{equation}

\begin{equation}
\label{eq:psi}
\Psi(w,h,P) = \sum_{k=0}^{l(w,h)-1}log(P_{(w,h)}^{k})+\sum_{k=l(w,h)}^{K-1}log(1-P_{(w,h)}^{k})
\end{equation}

\noindent
 $P_{(w,h)}^{k}$ is the softmax probability output of location ${(w,h)}$ for distance class ${k}$, and ${l(w,h)}$ is the target ordinal label that converted from the target depth using the spacing-increasing discretization method \cite{DORN}. Minimizing the $\mathcal{L}_{ordinal}$ will ensure the distance classification result of the prediction is close to the target label, and the model will result in a better output estimation. Note that both $\mathcal{L}_{L1}$ and $\mathcal{L}_{ordinal}$ are calculated with non-zero value pixels in ground truth target depth.


\begin{figure*}[ht]
\includegraphics[width=\textwidth,height=\textheight,keepaspectratio]{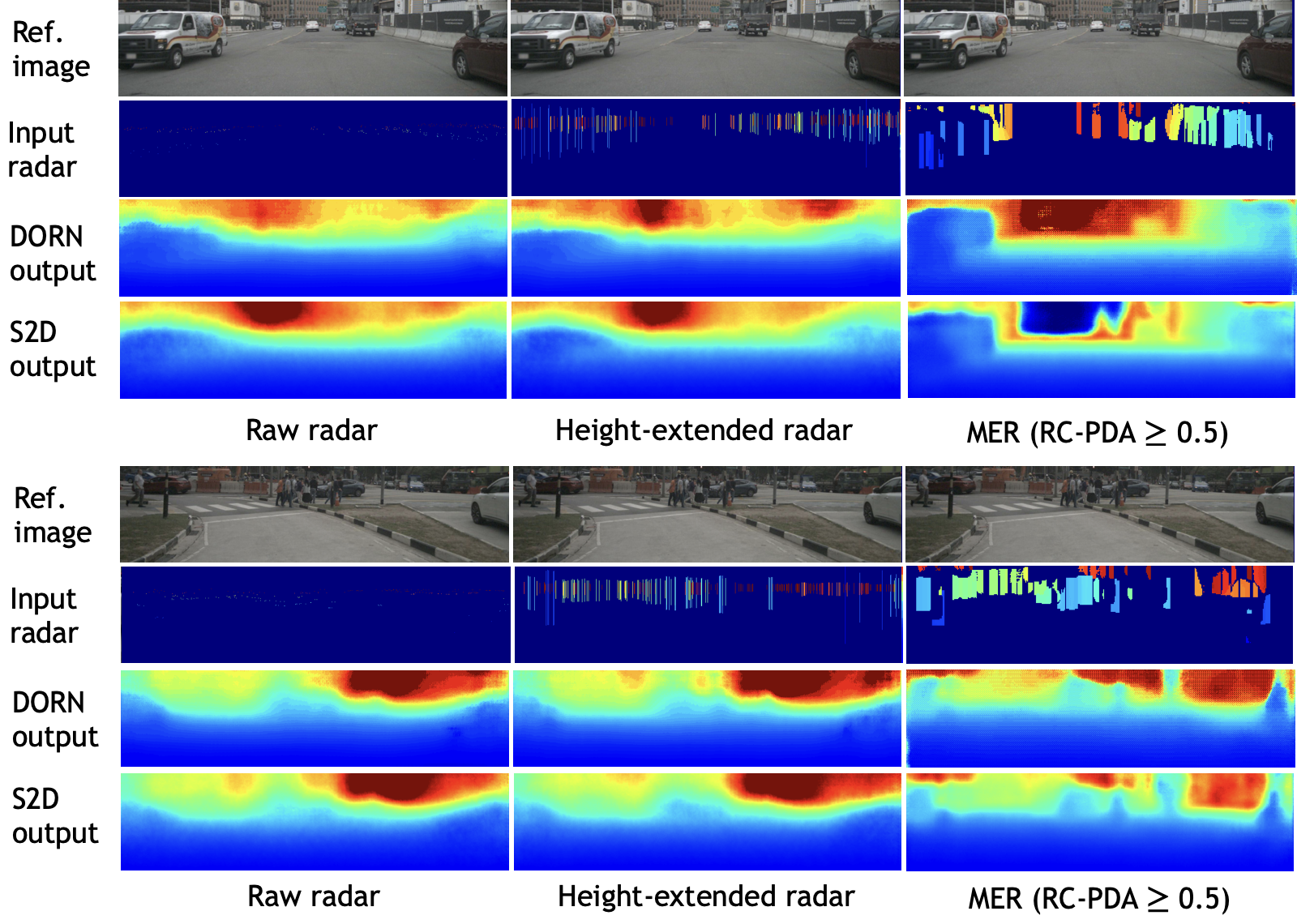}
  \caption{Qualitative comparison of results for radar inference experiments. 
  From left to right: Raw radar; Height-extended radar; MER as input signals. From top to bottom: reference camera image; input radar signal; output prediction from DORN$_{radar}$; output prediction from S2D$_{radar}$.}
  \label{fig:radar_inference}
\end{figure*}



\section{Experiments}
\label{sec:experiments}
\vspace{1mm}
\subsection{Dataset and Implementation}
\vspace{1mm}
\subsubsection{NuScenes Dataset}
\vspace{1mm}
We conduct our experiments on the nuScenes dataset~\cite{nuScenes}, the most commonly used dataset for integrating radar in both depth estimation and 3D object detection tasks. The nuScenes dataset is currently one of the most comprehensive multi-modal autonomous driving datasets consisting of 6 cameras, 5 FMCW radars, and a 32-beam Velodyne lidar. There are 1000 driving scenes captured in Boston and Singapore, and each scene contains roughly 40 manually synchronized samples from a 20s recording of driving. There are 850 scenes officially split into 700 training and 150 validation scenes. We use the front view data only, resulting in 28130 training and 6019 validation samples.

\subsubsection{Implementation Details}
\vspace{1mm}
All the models are run from the code released with the original papers and trained on a Tesla V100 GPU. To ease computation, the camera images, projected lidar depth, and radar depth are downsampled from the original shape of $900\times 1600$. For radar inference experiments, we downsample the input image to $450\times 800$ and further crop into the size of $350\times 800$ for both input and output since the upper region contributes no useful depth information. For radar supervision experiments, we use an input size of $350\times 800$ and an output size of $88\times 200$ because a model is easier to learn the overall shape on a smaller resolution of output prediction. The ground truth lidar depth for training radar inference experiments is densely interpolated with sparse lidar and camera images by the colorization method~\cite{colorized} as also used in~\cite{DORN_radar}. For both experiments, three variations of radar are used (raw, height-extended, and MER), and we accumulate raw and height-extended radar data with the current frame and the previous 4 frames. The height-extended radar extends each projected radar point to a height range of 0.25m to 2m, which is the same setting as in~\cite{DORN_radar}. We use the MER channel with RC-PDA $\geq 0.5$. The DORN$_{radar}$ and S2D$_{radar}$ are referred from~\cite{DORN_radar},~\cite{MDE_radar} respectively. Polynomial decay with an initial learning rate of 0.0001 and a power rate of 0.9 is applied as the learning strategy. The batch size is set to 8, and momentum and weight decay are set to 0.9 and 0.0005, respectively. The S2D$_{radar}$ and S2D$_{RGB}$ are trained with L1 loss while DORN$_{radar}$ uses ordinal loss. All the experiments are set to train for 30 epochs on the nuScenes official training splits, and test on the nuScenes official validation splits. The evaluation metrics used are the standard evaluation metrics also used in previous works and calculations for both experiments are based on the size of $350\times 800$ using ground truth sparse lidar with a maximum distance of either 50m or 80m.


\begin{figure*}[ht]
  \includegraphics[width=\textwidth,height=\textheight,keepaspectratio]{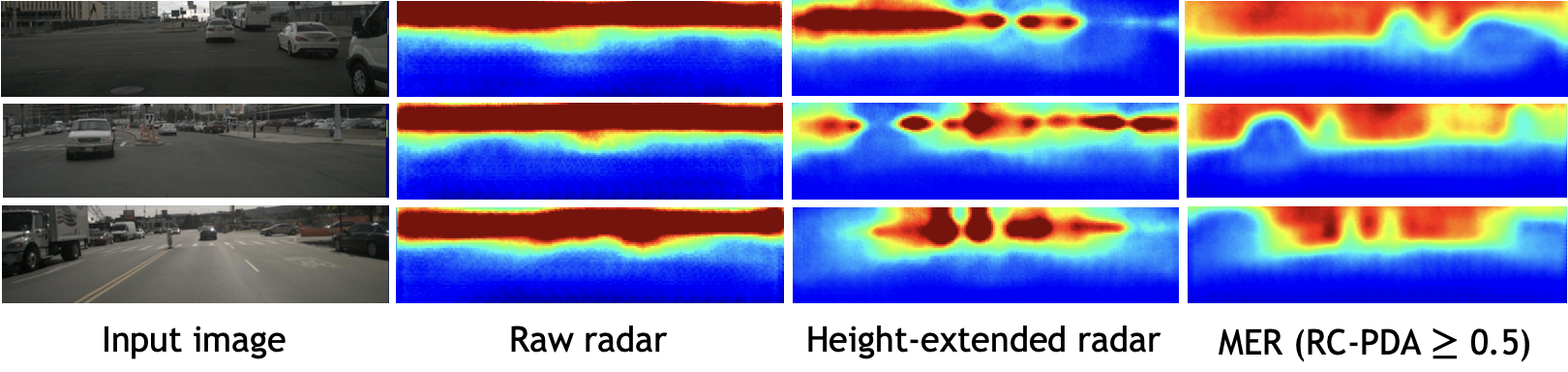}
  \caption{Qualitative comparison of results for radar supervision experiments.
  From left to right: input camera image; output prediction of the model supervised by raw radar; height-extended radar; MER. 
  }
  \label{fig:radar_supervision}
\end{figure*}


\subsection{Evaluation Metrics}

We evaluate the results with the following metrics:

• $\delta$$_n$: percentage of estimated pixels for which the relative
error is within a threshold:
$$
\delta_{n} = \frac{1}{WH} \left( y_{target}: max(\frac{y_{target}}{y_{pred}},\frac{y_{pred}}{y_{target}}) < 1.25^n \right )
$$

• RMSE: Root Mean Square Error.
$$
RMSE = \sqrt{\frac{1}{WH} \sum_{w=0}^{W-1}\sum_{h=0}^{H-1} \left \| y_{target}(w,h)-y_{pred}(w,h) \right \|^2_2 }
$$

• AbsRel: Mean Absolute Relative Error.
$$
AbsRel = \frac{1}{WH} \sum_{w=0}^{W-1}\sum_{h=0}^{H-1} \frac{\left| y_{pred}(w,h)-y_{target}(w,h) \right|}{\left| y_{pred}(w,h) \right|}
$$

\noindent
where ${(w,h)}$ is the pixel location, and $y_{pred}$ and $y_{target}$ are the output prediction of the model and target depth, respectively. Note that all the metrics are calculated based on non-zero value pixels in $y_{target}$.

\subsection{Radar Inference Experiments}
\vspace{1mm}
The idea of our radar inference experiments is to examine if a model is able to predict surroundings to a fair extent with only radar input and under lidar supervision during training. The results is shown in Table \ref{tab:radar_inference}, and the qualitative result of this experiment is visualized in Fig. \ref{fig:radar_inference}. The raw radar has the poorest performance among all since it is the most sparse one but still reaches around 0.7 for $\delta${$_1$}. Due to the supervision with a lidar signal, the model learns to generate a more general output as shown in the qualitative visualization. The results of both preprocessed radar signals are comparable to each other in metrics, but it shows in the qualitative results that the result using MER input could capture more detail compared to height-extended radar input. This indicates that increasing the intelligibility of radar data can improve performance. Among all settings, the S2D$_{radar}$ with MER has 0.8 for $\delta${$_1$} as the highest score in $\delta${$_1$}, which means 80\% of the predicted depth values is within 25\% of difference compared to the ground truth lidar. Since MER is preprocessed radar data generated by a neural network, it seems fair that this input shows the best performance. It is clear from both the metrics and the figures that the predictions based on radar input can detect the shape of surroundings and vehicles to a fair extent in raw and height-extended radar. Additionally, the model can gain much performance with an advanced preprocessing method that can adequately densify and filter raw radar data.

\subsection{Radar Supervision Experiments}
\vspace{1mm}
The idea of radar supervision experiments is to investigate how good the performance of a monocular depth estimation model can be by taking camera images as input and training under the supervision of radar data instead of ground truth lidar. Note that ground truth lidar is used only for calculating evaluation metrics and not as an input signal or supervision target during training.
We conduct this experiment based on the S2D$_{RGB}$ model, and the evaluation results are shown in Table \ref{tab:radar_supervision} while the qualitative result is shown in Fig. \ref{fig:radar_supervision}. The result shows the same trend compared to radar inference experiments, that both preprocessed radars outperform raw data, indicating that the overall performance improves with proper preprocessing of radar data. Additionally, both height-extended radar and MER can achieve 0.6 in $\delta${$_1$}, which is about 70\% of the performance compared to the result of the baseline model trained with sparse lidar supervision. The qualitative result also confirms that the performance of the model trained with preprocessed radar improves significantly. Most of the vehicles and obstacles in the middle or at the edge are nicely detected with MER supervision compared to the supervision of raw data and height-extended radar in Fig. \ref{fig:radar_supervision}. This reveals the potential of using height-extended radar or MER as a supervision target in depth estimation tasks and reducing the demand for lidar data.



\section{Conclusion}
\label{sec:conclusion}
\vspace{1mm}
In this paper, we conducted radar inference and supervision experiments to show how much depth information radar can contribute to depth estimation tasks. Our quantitative results from the inference experiment show that the inference capability of radar data is fair and agrees with the results from prior works after preprocessing but is still limited. However, the radar supervision experiment shows the opposite trend. The supervision experiment revealed that a monocular depth estimation model could predict to a reasonable extent under the supervision of preprocessed radar. This result indicates that radar can contribute more depth information as a supervision signal after proper preprocessing, which gives a potential opportunity to treat radar as an additional supervision target and ease the usage of lidar. Additionally, there is also a possibility to see radar as a domain adaptation target on a well-trained monocular depth estimation. To the best of our knowledge, this is the first paper to conduct such experiments to investigate the intrinsic depth information of radar data.



\begin{biography}

Chen-Chou Lo received an M.Sc. degree in electrical engineering from National Central University, Taiwan, in 2014. He has been working toward a Ph.D. degree in the Department of Electrical Engineering at KU Leuven, Belgium, since 2019. From 2018 to 2019, he worked at Academia Sinica as a research assistant in Taiwan, and he joined the EAVISE research group in KU Leuven in 2019. His current research focuses on the development of fusing radar and camera data on depth estimation and 3D object detection in an autonomous driving scenario.

Patrick Vandewalle received a M.Sc. degree in electrical engineering from KU Leuven, Belgium, in 2001, and a Ph.D. degree from EPFL, Switzerland, in 2006. From 2007 to 2018, he worked at Philips Research, The Netherlands, as a senior research scientist. He is now an associate professor at KU Leuven, Belgium. His current research in the EAVISE research group focuses on 3D processing, reconstruction, computer vision and AR/VR.

\end{biography}


\small
\begin{thebibliography}{9}
\bibitem{PSMNet} Jia-Ren Chang and Yong-Sheng Chen, Pyramid Stereo Matching Network, Proceedings of the IEEE/CVF Conference on Computer Vision and Pattern Recognition (CVPR), 2018.
\bibitem{GA-Net} Feihu Zhang, Victor Prisacariu, Ruigang Yang, and Philip H.S. Torr, GA-Net: Guided Aggregation Net for End-To-End Stereo Matching, Proceedings of the IEEE/CVF Conference on Computer Vision and Pattern Recognition (CVPR), 2019.
\bibitem{eigen_1} David Eigen, Christian Puhrsch, and Rob Fergus, Depth Map Prediction from a Single Image using a Multi-Scale Deep Network, Advances in Neural Information Processing Systems (NeurIPS), pg. 2366-2374. 2014.
\bibitem{eigen_2} David Eigen and Rob Fergus, Predicting Depth, Surface Normals and Semantic Labels with a Common Multi-scale Convolutional Architecture, IEEE International Conference on Computer Vision (ICCV), pg. 2650-2658. 2015.
\bibitem{DORN} Huan Fu, Mingming Gong, Chaohui Wang, Kayhan Batmanghelich, and Dacheng Tao, Deep Ordinal Regression Network for Monocular Depth Estimation, Proceedings of the IEEE/CVF Conference on Computer Vision and Pattern Recognition, pg. 2002-2011. 2018.
\bibitem{bts} Jin Han Lee, Myung-Kyu Han, Dong Wook Ko, and Il Hong Suh, From big to small: Multi-scale local planar guidance for monocular depth estimation, arXiv preprint:1907.10326. 2019.
\bibitem{MSG-CHN} Ang Li, Zejian Yuan, Yonggen Ling, Wanchao Chi, Shenghao Zhang, and Chong Zhang, A Multi-Scale Guided Cascade Hourglass Network for Depth Completion, Proceedings of the IEEE/CVF Winter Conference on Applications of Computer Vision (WACV), 2020.
\bibitem{guide_uncertainty} Wouter Van Gansbeke, Davy Neven, Bert De Brabandere, and Luc Van Gool, Sparse and Noisy LiDAR Completion with RGB Guidance and Uncertainty, Proceedings of the 16th International Conference on Machine Vision Applications (MVA), pg. 1-6. 2019. 
\bibitem{s2d} Fangchang Ma and Sertac Karaman, Sparse-to-Dense: Depth Prediction from Sparse Depth Samples and a Single Image, Proceedings of the IEEE International Conference on Robotics and Automation (ICRA), 2018.
\bibitem{MDE_radar} Juan-Ting Lin, Dengxin Dai, and Luc Van Gool, Depth Estimation from Monocular Images and Sparse Radar Data, Proceedings of the IEEE International Conference on Intelligent Robots and Systems (IROS), 2020.
\bibitem{DORN_radar} Chen-Chou Lo and Patrick Vandewalle, Depth Estimation From Monocular Images And Sparse Radar Using Deep Ordinal Regression Network, Proceedings of the IEEE International Conference on Image Processing (ICIP), pg. 3343-3347. 2021. 
\bibitem{RC_PDA} Yunfei Long, Daniel Morris, Xiaoming Liu, Marcos Castro, Punarjay Chakravarty, and Praveen Narayanan, Radar-Camera Pixel Depth Association for Depth Completion, Proceedings of the IEEE/CVF Conference on Computer Vision and Pattern Recognition (CVPR), pg. 12507-12516 2021.
\bibitem{MDE_multitask} Wei-Yu Lee, Ljubomir Jovanov, and Wilfried Philips, Semantic-Guided Radar-Vision Fusion for Depth Estimation and Object Detection, Proceedings of the 32th British Machine Vision Conference (BMVC), 2021.
\bibitem{RVMDE} Muhamamd Ishfaq Hussain, Muhammad Aasim Rafique, and Moongu Jeon, RVMDE: Radar Validated Monocular Depth Estimation for Robotics, arxiv, 2021.
\bibitem{nuScenes} Holger Caesar, Varun Bankiti, Alex H. Lang, Sourabh Vora, Venice Erin Liong, Qiang Xu, Anush Krishnan, Yu Pan, Giancarlo Baldan, and Oscar Beijbom, nuScenes: A Multimodal Dataset for Autonomous Driving, Proceedings of the IEEE/CVF Conference on Computer Vision and Pattern Recognition (CVPR), pg. 11618-11628. 2020.
\bibitem{BANet} Shubhra Aich, Jean Marie Uwabeza Vianney, Md Amirul Islam, Mannat Kaur, and Bingbing Liu, Bidirectional Attention Network for Monocular Depth Estimation, Proceedings of the IEEE International Conference on Robotics and Automation (ICRA), 2021.
\bibitem{zhu2020edge} Shengjie Zhu, Garrick Brazil, and Xiaoming Liu, The edge of depth: Explicit constraints between segmentation and depth, Proceedings of the IEEE/CVF Conference on Computer Vision and Pattern Recognition (CVPR), pg. 13116-13125. 2020.
\bibitem{monodepth1} Clément Godard, Oisin Mac Aodha, and Gabriel J. Brostow, Unsupervised Monocular Depth Estimation with Left-Right Consistency, Proceedings of the IEEE/CVF Conference on Computer Vision and Pattern Recognition (CVPR), 2017.
\bibitem{monodepth2} Clément Godard, Oisin Mac Aodha, Michael Firman, and Gabriel Brostow, Digging into Self-Supervised Monocular Depth Prediction, Proceedings of the International Conference on Computer Vision (ICCV), 2019.
\bibitem{SparseAD} Maximilian Jaritz, Raoul de Charette, Emilie Wirbel, Xavier Perrotton, and Fawzi Nashashibi, Sparse and Dense Data with CNNs: Depth Completion and Semantic Segmentation, Proceedings of the International Conference on 3D Vision (3DV), pg. 52-60. 2018.
\bibitem{ResNet} Kaiming He, Xiangyu Zhangg, Shaoqing Ren, and Jian Sun, Deep Residual Learning for Image Recognition, IEEE Conference on Computer Vision and Pattern Recognition (CVPR), pg. 770-778. 2016.
\bibitem{ASPP} Liang-Chieh Chen, George Papandreou, Iasonas Kokkinos, Kevin Murphy, and Alan L. Yuille, DeepLab: Semantic Image Segmentation with Deep Convolutional Nets, Atrous Convolution, and Fully Connected CRFs, IEEE Transactions on Pattern Analysis and Machine Intelligence, pg. 834-848. 2018.
\bibitem{colorized} Anat Levin, Dani Lischinski, and Yair Weiss, Colorization using optimization, ACM Trans. Graph., pg. 689-694. 2004.

\end{thebibliography}
\end{document}